\documentclass{svproc}
\usepackage{url}
\usepackage{amssymb}
\usepackage{subcaption}
\usepackage[fleqn]{amsmath}
\usepackage{graphicx}
\usepackage{multirow}
\usepackage{floatrow}
\usepackage{amsfonts}
\usepackage{hyperref}
\usepackage{svg}
\usepackage[ruled]{algorithm2e}

\begin{document}
\mainmatter              
\title{Task Space Control of Hydraulic Construction Machines using Reinforcement Learning}
\titlerunning{Task Space Control of Hydraulic Construction Machines using RL}  
%
\author{Hyung Joo Lee\inst{1} \and Sigrid Brell-Cokcan\inst{1}}
\authorrunning{Hyung Joo Lee and Sigrid Brell-Cokcan} 
%
\tocauthor{Hyung Joo Lee and Sigrid Brell-Cokcan}
\institute{Chair of Individualized Production (IP), RWTH Aachen University, Campus-Boulevard 30, 52074 Aachen, Germany,\\
\email{{lee, brell-cokcan}@ip.rwth-aachen.de}}
\maketitle              
\begin{abstract}
Teleoperation is vital in the construction industry, allowing safe machine manipulation from a distance. However, controlling machines at a joint level requires extensive training due to their complex degrees of freedom. Task space control offers intuitive maneuvering, but precise control often requires dynamic models, posing challenges for hydraulic machines. To address this, we use a data-driven actuator model to capture machine dynamics in real-world operations. By integrating this model into simulation and reinforcement learning, an optimal control policy for task space control is obtained. Experiments with Brokk 170 validate the framework, comparing it to a well-known Jacobian-based approach.
\end{abstract}
\section{Introduction}
Unlike the manufacturing industry, the construction sector poses unique challenges to robots due to its dynamic and diverse characteristics \cite{lee1}. Construction sites constantly undergo changes throughout different stages of development, each presenting distinct hazardous conditions for human workers \cite{lee4}. To mitigate these risks, teleoperation has become a crucial component of construction machinery in today's construction industry \cite{brell}. However, the complex nature of these machines, which often have multiple degrees of freedom (DoF) and require individual levers for remote control at the joint level, necessitates extensive operator training. Even experienced operators may need several months of training to effectively coordinate multiple joints and achieve the desired end-effector or tool motion. As a result, productivity decreases, local accuracy is reduced, and work efficiency is compromised.

Despite developing intelligent robotic systems equipped with advanced control and planning algorithms to address the challenges of unstructured construction sites and provide automation benefits, their autonomy is often constrained to highly unstructured environments. The reason behind this limitation lies in the impact of incomplete and inaccurate information regarding unfamiliar objects or unforeseen situations, which can significantly affect the decision-making process of robots and undermine their capacity for autonomous operation. To surpass the limitations faced by autonomous robots in unstructured environments, ongoing research aims to enhance operator capabilities through teleoperation systems that integrate automation techniques, such as virtual fixtures \cite{lee2} or task space control \cite{Khatib}. 

Task space control enables the robot to manipulate its actions and interactions with the environment intuitively and efficiently, considering its multi-DoF nature. When the robot model is accurately known, task space control is well-established and offers various alternatives for resolution, including resolved-motion rate control, resolved-acceleration control, and force-based control \cite{Nakanishi}. However, constructing a dynamic model of a construction machine presents significant challenges due to various factors. The intricate mechanical structure of the machine and the complex interactions among its numerous components, such as hydraulic actuators, linkages, and sensors, make it difficult to model the system's dynamics accurately. Furthermore, the presence of nonlinearities like friction, backlash, and hysteresis adds further complexity to the modeling process, requiring advanced techniques to address these effects effectively \cite{lee3}.

To address these challenges and enable effective task space control, we propose a framework utilizing a data-driven approach based on reinforcement learning (RL). In this approach, an agent is trained in a dynamic simulator and then directly deployed in the real-world environment. It is important to note that during the initial training phase, the agent may exhibit unpredictable behavior, which can raise safety concerns, particularly when dealing with heavy-duty machines. To overcome this issue, the task space control policy is learned through simulation and can be seamlessly applied to the real machine without requiring any parameter adjustments or post-processing. To the best of our knowledge, the application of general learning approaches to achieve task space control with a full scale of construction machines has not been adequately addressed.

The structure of this paper is as follows: We provide a brief description of the construction machine, Brokk 170, which serves as the basis for evaluating our proposed approach. Next, we introduce the RL framework, where the agent learns the task space control policy. To bridge the gap between simulation and the real-world, we employ a data-driven actuator model within the RL framework. Furthermore, we present a detailed explanation of how the agent learns the task space control policy using RL techniques by integrating the data-driven actuator model. Finally, we demonstrate the effectiveness of our proposed framework by deploying the trained agent on a real machine, showcasing its capabilities and potential benefits.

\section{System Description}

\begin{figure*}[!t] 
\begin{center}
\includegraphics[scale=0.45]{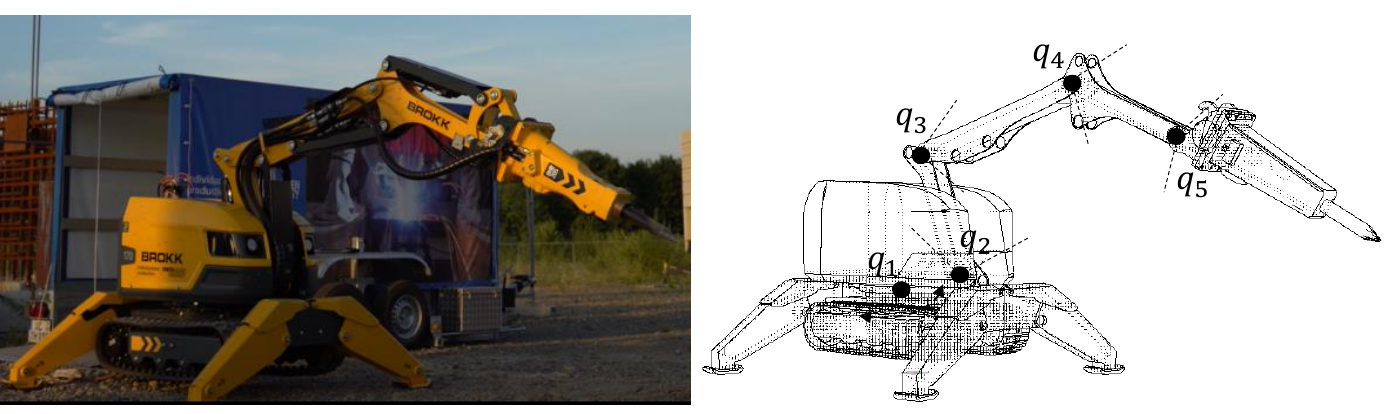}
\caption{Brokk 170 (Left) and the geometric representation in the joint space (Right).}
\label{brokk170}
\end{center}
\end{figure*}

This section presents an overview of the hardware configuration of the Brokk 170 construction machinery, which is utilized to evaluate the research work. The Brokk 170 is a mobile construction machine employed for various tasks such as drilling, demolition, and cutting, depending on the specific end-effector tool attached. The machine consists of linkage and actuator systems that are connected to a mobile base, while the base itself is considered to be locally fixed throughout the study. In modern construction machinery like the Brokk 170, several electronic components are integrated into the system, including sensors. These components are connected to the machine control unit (MCU) through a Controller Area Network (CAN) bus. When the operator sends commands using a control device such as joysticks, specific bus messages are generated and translated into Pulse-Width Modulation (PWM) signals with designated voltage levels. These PWM signals are then transmitted to specific power electronics responsible for driving the machine's mechanical components. Each joint of the Brokk 170 is equipped with an absolute multiturn encoder IFM RM9000, which enables accurate tracking of the joint motion. A controller IFM CR711S and an embedded PC Jetson AGX Xavier are utilized to facilitate communication between the host PC and the MCU. The MCU, which is responsible for controlling the machine's valve system, operates at a frequency of 20Hz as specified by the manufacturer. In this work, instead of directly producing the PWM duty cycle from the joystick movement, the different joystick movements are converted into task space goals $\textbf{x}=(\dot{x}, \dot{y}, \dot{z})$. These specified task space goals are then translated into corresponding PWM values for each joint in the host PC and wirelessly transmitted to the MCU through the controller, which has the CAN bus interface.

Brokk 170 is a hydraulic serial link manipulator consisting of five revolute joints. It is important to note that the second joint, denoted as $q_2$, is physically connected to $q_3$ (refer to Fig. \ref{brokk170}). According to the manufacturer's specifications, $q_2$ is intended exclusively for adjusting the reach of the manipulator. To ensure the proper functioning of the machine, the manufacturer has established a communication protocol that restricts the rotation of $q_2$ in conjunction with other joints. Therefore, in this study, the focus is solely on implementing task-space control using joints $q_1, q_3, q_4,$ and $q_5$ while disregarding the involvement of $q_2$ in the control process.

\begin{figure}[!t]
\begin{minipage}{0.5\textwidth}
  \begin{subfigure}{\linewidth}
  \includegraphics[width=0.95\linewidth]{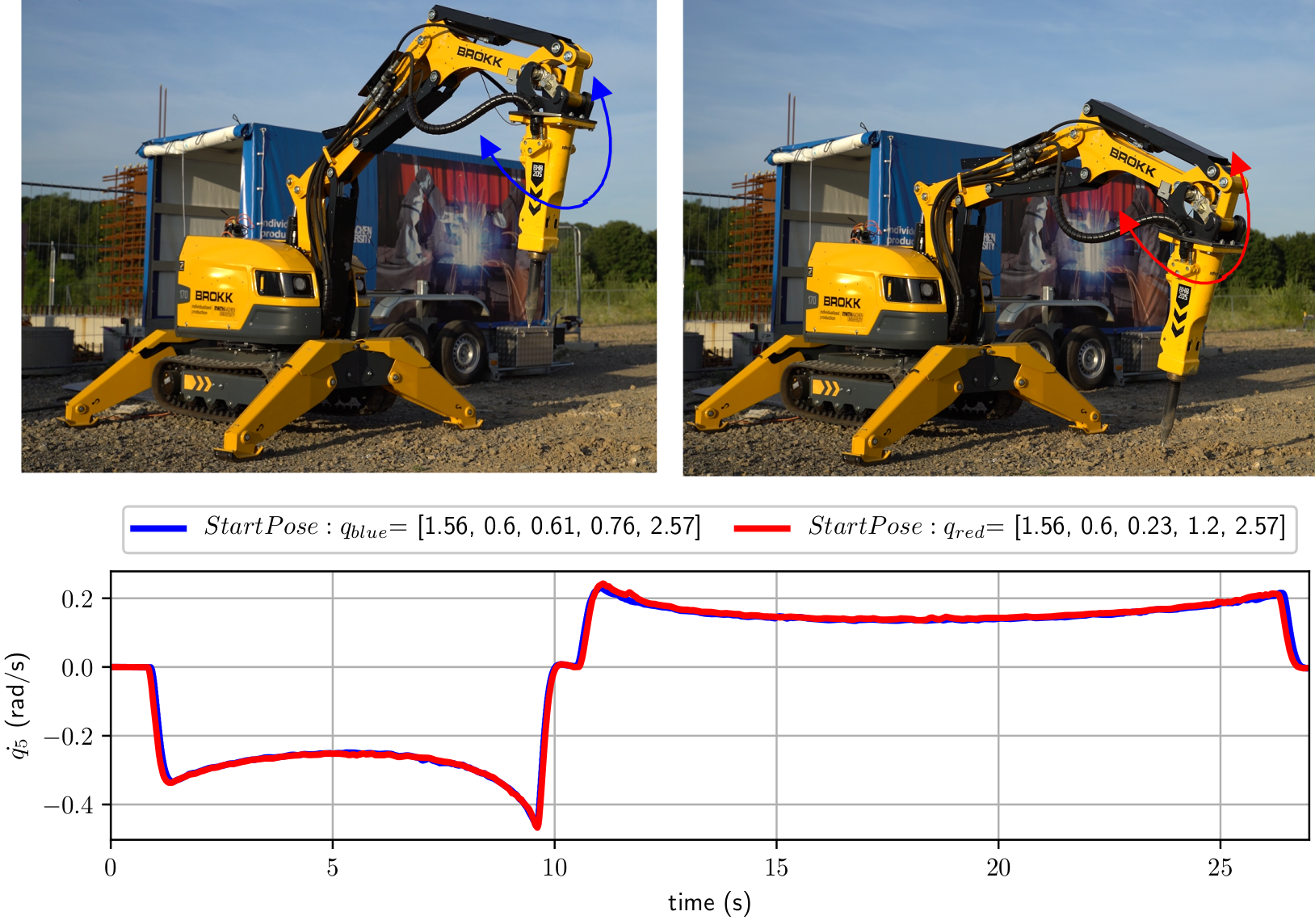}  
  \caption{Velocity profiles of $q_5$, when commanded $pwm_5$ values of 50 and 205 at different joint configurations of the machine.}
  \end{subfigure}
\end{minipage}%
\begin{minipage}{0.5\textwidth}
  \begin{subfigure}{\linewidth}
  \includegraphics[width=.95\linewidth]{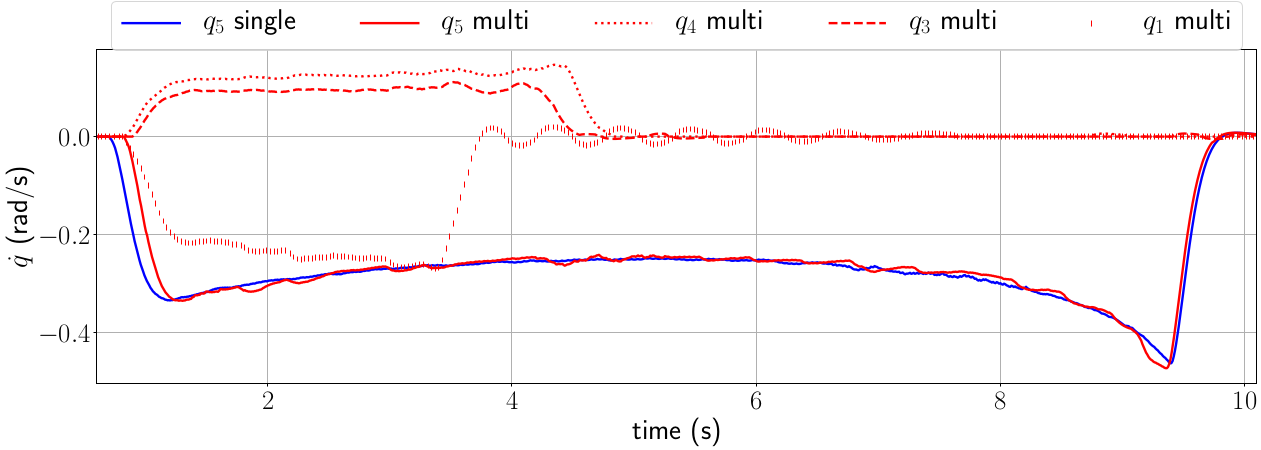} 
    \caption{Velocity profiles of $q_5$ and the corresponding $pwm_5$ normalized in [0,1]. }
  \includegraphics[width=.95\linewidth]{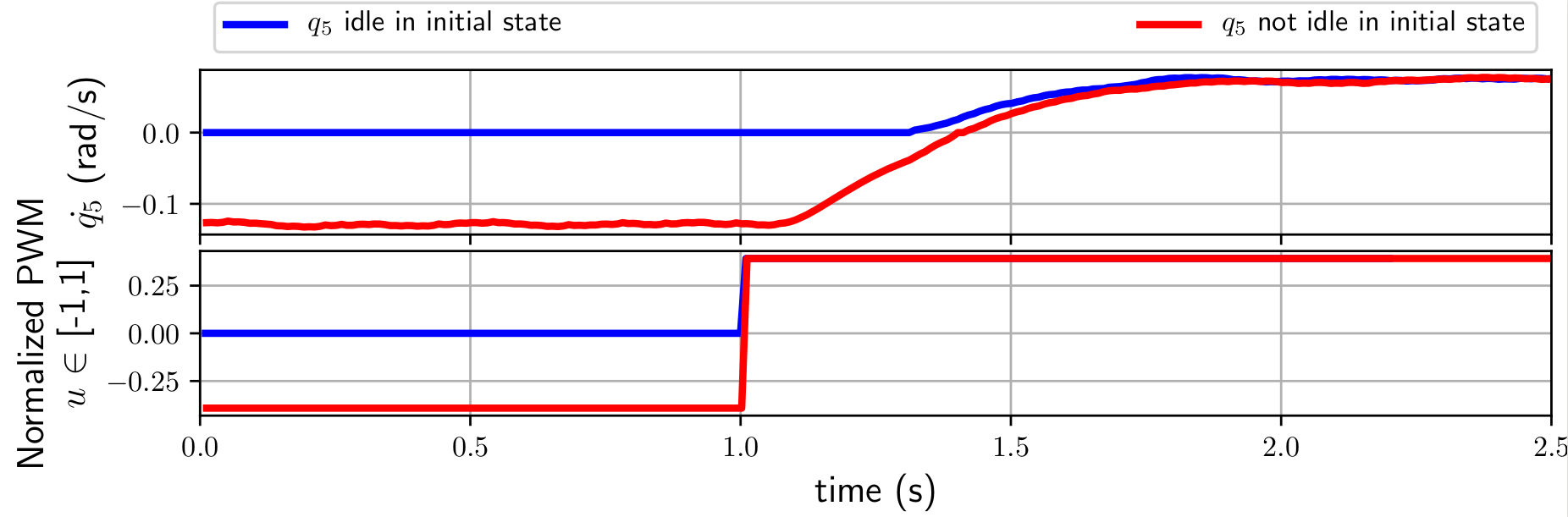}  
  \caption{Velocity profiles of $q_5$ were compared when actuated individually (blue) and simultaneously with multiple actuators (red).}
  \end{subfigure}
\end{minipage}
\caption{Primary experiment tests results.}
\label{primary}
\end{figure}

\section{Method}
Reinforcement learning (RL) revolves around the iterative process of collecting data through trial and error, automatically adjusting the control policy to optimize a cost or reward function that represents the task at hand \cite{sutto}. This approach offers a fully automated means of optimizing the control policy, encompassing everything from sensor readings to low-level control signals. It provides a flexible framework for discovering and refining skills to solve complex tasks. However, RL typically demands extensive interaction time with the system to learn intricate skills, often requiring hours or even days of real-time execution \cite{levine}. Moreover, during the training phase, machines may exhibit sudden and unpredictable behavior, raising safety concerns, particularly when dealing with heavy-duty tasks.

To effectively train construction machines in handling complex tasks, our proposed methodology leverages advanced physics simulation technology. This involves training the machines in a simulated environment and transferring the acquired skills to human operators for real-world scenarios. By employing this approach, we mitigate the risk of unpredictable machine behavior during initial training stages and reduce the need for extensive real-world training data. However, the success of this methodology hinges on effectively bridging the reality gap between the simulated and real-world systems. To address this challenge, we incorporate the concept of a data-driven actuator model \cite{Hwangbo}, which helps reconcile the disparities in system dynamics between the simulation and real-world environments.

\subsection{Data-driven Actuator Model}

\subsection{Preliminary Tests}
The selection of input and output sets in the network significantly influences the performance of predictions. In this study, we conduct experimental investigations to understand the factors that affect the mapping behavior between the control input (PWM) and joint velocity. This analysis aims to determine the appropriate input and output settings for achieving optimal results. It is important to note that when applying the presented approach to a different machine, distinct input and output sets for the neural network model are required, as the outcomes can vary depending on the machine type and manufacturer. Consequently, these initial experiments also serve as a fundamental guideline for identifying the machine's relevant characteristics in relation to the proposed data-driven control approach, which is typically not provided by manufacturers.

\subsubsection{Actuation Range}
The hydraulic construction machine employed in this work includes five joints. Each joint has a hydraulic cylinder that moves in a linear motion, where the cylinder’s linear movement is subsequently transformed into the joint’s rotation. This relation between the cylinder and joint motion can be described as

\begin{equation}\label{eq_1}
\omega_i(q_i) = \dfrac{\nu_i}{r_i (q_i)}
\end{equation}
where $i$ indicates the joint number, $\omega$ is the angular velocity of the joint, $\nu$ is the linear velocity of the cylinder, and $r$ is the distance between the axis of the cylinder and the pivot point. The axis rotates for the constant linear speed of the cylinder $\nu$. As a result, the distance $r$ changes, resulting in a changing angular velocity, as one can see in Fig. \ref{primary}a.\\

\subsubsection{Cylinder Movement Direction}
The manipulator is inherently affected by gravity. If the joint rotates towards gravity, the according rotational speed is typically larger than rotating in the opposite direction. In this experiment, $q_5$ was rotated with the same PWM signal but a different sign (i.e., $u_5$= 80 and $u_5$= 175, respectively) throughout the actuation range. Here $u_5$ $\in$ [0,127] corresponds to the rotation into the positive direction and $u_5$ $\in$ [255,128] into the negative direction. The corresponding rotational speed $\dot{q_5}$ is reported in Fig. \ref{primary}a. We conducted this test multiple times from different postures of the machine, as shown in Fig. \ref{primary}a. As expected, the velocity was larger when rotating towards gravity with the minimum value around -0.47 $\frac{rad}{s}$. In contrast, in the other instance, the maximum velocity was around 0.24 $\frac{rad}{s}$. The different joint configurations did not have any impact on the velocity profile.\\

\subsubsection{Coupled Dynamics}
In general, coupled dynamics occur in hydraulic actuators if the fluid pump cannot deliver the requested fluid flow. In other words, the data collection process must be designed to sufficiently capture the coupled dynamics by simultaneously actuating multiple cylinders in the presence of the coupled dynamics. Thus, this test was conducted to analyze whether the system's behavior changes when numerous joints are actuated. Through this work, the maximum value for PWM is restricted to about 65 $\%$, i.e., 80 and 175, respectively. The reason for this is that the manipulator's impulsive movement with multiple joints actuation at full speeds can possibly cause the overturning of the machine or undesired collision with the environment. Thus, for the test, $q_5$ was rotated along its actuator range with $u_5$= 80. At the second time, the $q_1$, $q_3$ and $q_4$ were also simultaneously rotated with $u_1$= 175, $u_3$= 175 and $u_4$= 80, respectively, while $u_5$= 80 was applied to $q_5$ like the first test. The result in Fig. \ref{primary}b shows that the difference is marginal and neglectable. \\

\subsubsection{Past Movement}
BROKK 170 used in this work has a slow system reaction, as shown in Fig \ref{primary}c. As a result, the system’s previous states often influence the next state, i.e., even if the same PWM signal is transmitted to the same system, the resulting velocity profile varies depending on the system’s prior and present velocity. Fig. \ref{primary}c clearly shows this behavior. Here, the same PWM signal is sent to $q_5$ at 1$s$. In the first test, $q_5$ starts from its idle state, not moving. The response time is roughly 400$ms$ in this case. When $q_5$ is already moving, the response time is faster with approximately 100$ms$. As this result clearly shows, the past states play a vital role in accurately describing the correlation between $u$ and $\dot{q}$, particularly before $\dot{q}$ converges.

\begin{figure}[!t] 
\centering
\includegraphics[scale=0.6]{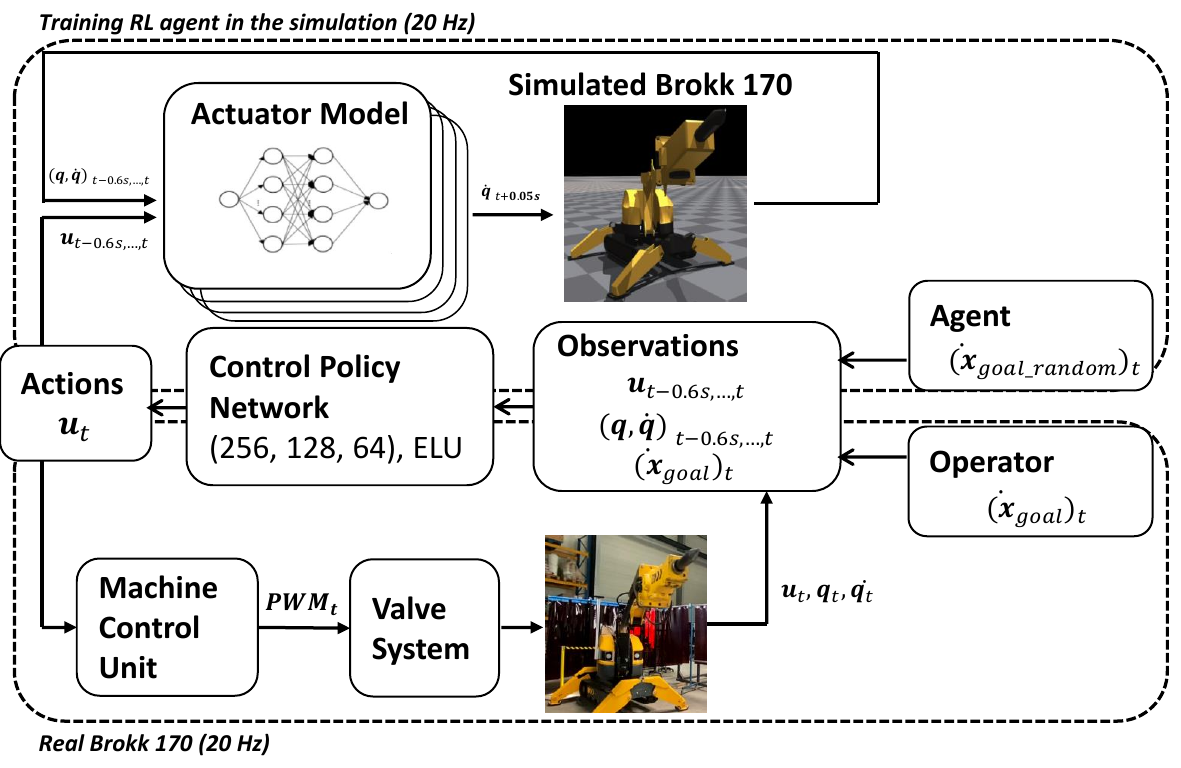}
\caption{Overview of proposed system.}
\label {overview}
\end{figure}

\begin{table}[t!]
\centering
\caption{Inputs/Outputs of the data-driven actuator model}
\label{table_example}
\begin{tabular}{ll}
\textbf{Inputs}                 &   \\ \hline
Joint Positions       & ${q_i^{t-0.6s}}$,${q_i^{t-0.55s}}$, $\cdots$, ${q_i^{t}}$ \\
Joint Velocities      &   $\dot{q_i}^{t-0.6s}$,$\dot{q_i}^{t-0.55s}$, $\cdots$, $\dot{q_i}^{t}$ \\
Control Signals      &   $\dot{u_i}^{t-0.6s}$,$\dot{u_i}^{t-0.55s}$, $\cdots$, $\dot{q_i}^{t}$ \\
\textbf{Output}                &   \\ \hline
Joint Velocity                   & $\dot{q_i}^{t+0.05s}$\\
\end{tabular}
\end{table}

\subsection{Methods of Training}
Based on the primary experiment results from the previous section, the set of network input and output is defined, as listed in Table \ref{table_example}. To collect the training data that captures the actuator’s behavior in different directions and speeds, PWM signals are generated in a sine waveform. Here, the sine wave’s frequency and the amplitudes are randomly modified. The generated PWM signals are applied until the corresponding joint reaches its minimum or maximum, where the joint then returns back to the home configuration. Then the next sine wave is generated. The data collected during the return is not included in the training data. 

We collect the dataset during 1$h$ for each actuator separately at 20 Hz (5$h$ for all the actuators) as the coupling dynamics are found to be insignificant (see Fig. \ref{primary}b). By collecting the training data separately for each actuator, we can reduce the risk of collision between the machine and the environment, as the other actuators can be moved to a safe configuration and held still while obtaining the data from one actuator. The generated PWM signals are set to be larger than the dead zone to partially compensate for the dead zone, estimated to be roughly 7$\%$ of the maximum. Also, the PWM duty cycle is limited to 65$\%$ of its maximum due to safety. 

The nonlinear relationship between the given PWM and
the resulting joint velocity is modeled using a multi-layer
perceptron (MLP), which consists of two hidden layers with
32 units, a ReLu activation function, and a linear output
layer with a sigmoid activation function. All the input and
output data are normalized. The model is trained using
the mean squared error (MSE) loss function and Adam
optimizer, where the loss converges after around 0.75$h$ for each model.



\subsection{Learning the task-space control policy}

\begin{figure}[!t] 
\centering
\includegraphics[scale=0.38]{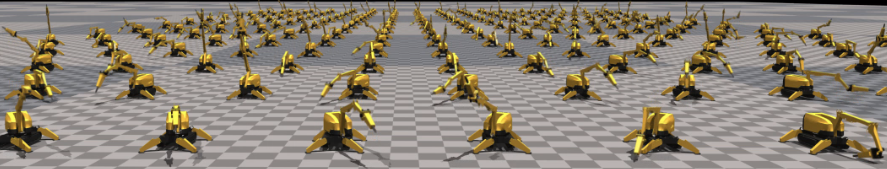}
\caption{Parallel simulation environment with 256 agents to train a control policy for task space control}
\label {isaac}
\end{figure}
The Brokk 170 robot utilized in this study possesses a total of five degrees of freedom (DoF), out of which four DoF are employed for task space control, as mentioned in the previous section. In this work, the robot's base is considered to be fixed. Consequently, the task space can be defined as $\textbf{x}=(x,y,z, \theta)$, where $\textbf{x}$ represents the position and orientation in the task space. However, when tracking a goal in the task space with the joint configuration $\textbf{q} \in \mathbb{R}^4$, the robot may encounter challenges in accurately following the goal. Therefore, in this study, we do not track the orientation component $\theta$ in order to demonstrate the performance of the implemented task space controller. By omitting the orientation tracking, we can focus on evaluating the performance of the task space control without additional errors arising from the limited robot configuration and kinematics.

To address the nonlinear dynamics of the actuators and achieve optimal control inputs, we employ RL, as it allows for learning through trial and error. In our approach, we integrate a real-world data-driven actuator model into the RL environment, enabling the RL agent to effectively handle input delays and nonlinear dynamics, thereby optimizing the coordination of the joints. 

\subsubsection{States and Observation}
At every time step $t$, the agent is provided with an observation that encompasses various details about the current state of the environment. These details include information like the present and target velocity of the end-effector, as well as the requisite data for the actuator network, see Fig. \ref{overview}. It is assumed that the kinematics of Brokk 170, specifically the joint position and link lengths, are already known. Therefore, utilizing only the joint values is sufficient for determining the position and velocity of the end-effector through forward kinematics. 

To expedite the training process, all observations and actions are standardized by applying a normalization technique using mean and standard deviation. In practical scenarios, the measured values of joint angles and the end-effector's corresponding position and velocity are susceptible to noise generated by machine vibrations. To account for this factor, we introduce uniformly sampled white noise into the observations. This noise is controlled by a scaling factor, maintaining a consistent maximum amplitude of 5\% throughout the entire episode.

\subsubsection{Rewards}

A reward function is utilized to guide the learning process and encourage the desired behavior in the control policy. Typically, this function incorporates penalties for actions that lead to undesired behaviors, such as collisions with the environment (in this case, the ground). However, considering the nature of construction machines that are intended to manipulate the environment for construction purposes, no additional penalty is implemented in this context. The reward function is designed such that the agent utilizes joints $q_{1,3,4,5}$ to control the end-effector velocity $v_t^{ee}$ according to the desired velocity $v_t^d$:

\begin{equation}
\begin{gathered}
    r_t= 1/(1+\lVert v_t^d - v_t^{ee} \rVert_2)   
\end{gathered}
\end{equation}

At every time step $t$, the reward $r_t$ is calculated and accumulated throughout each episode. This approach ensures that the agent is inherently motivated to promptly track the desired velocity, aiming to maximize the overall reward.

\begin{figure}[!t]
\centering
  \begin{subfigure}{\linewidth}
    \centering
    \includegraphics[scale=0.53]{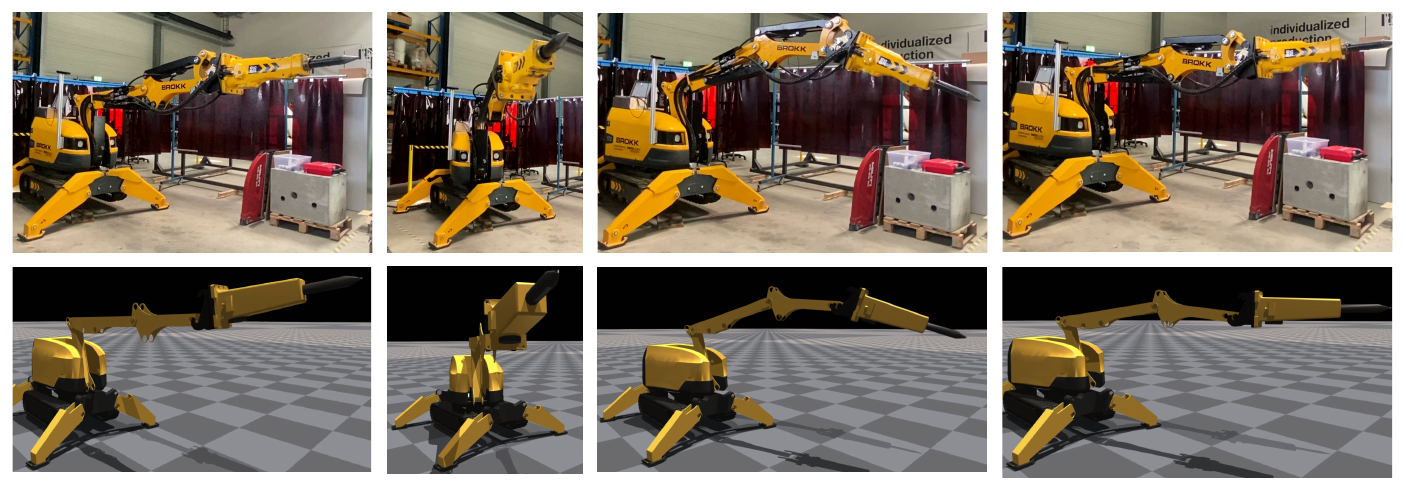}
    \caption{Snapshots from the experiments, where the same sequence of control signals are applied to the real machine (top) and the simulation (bottom) }
  \end{subfigure}
  
\begin{minipage}{0.48\textwidth}
  \centering
  \begin{subfigure}{\linewidth}
    \centering
    \includegraphics[scale=0.32]{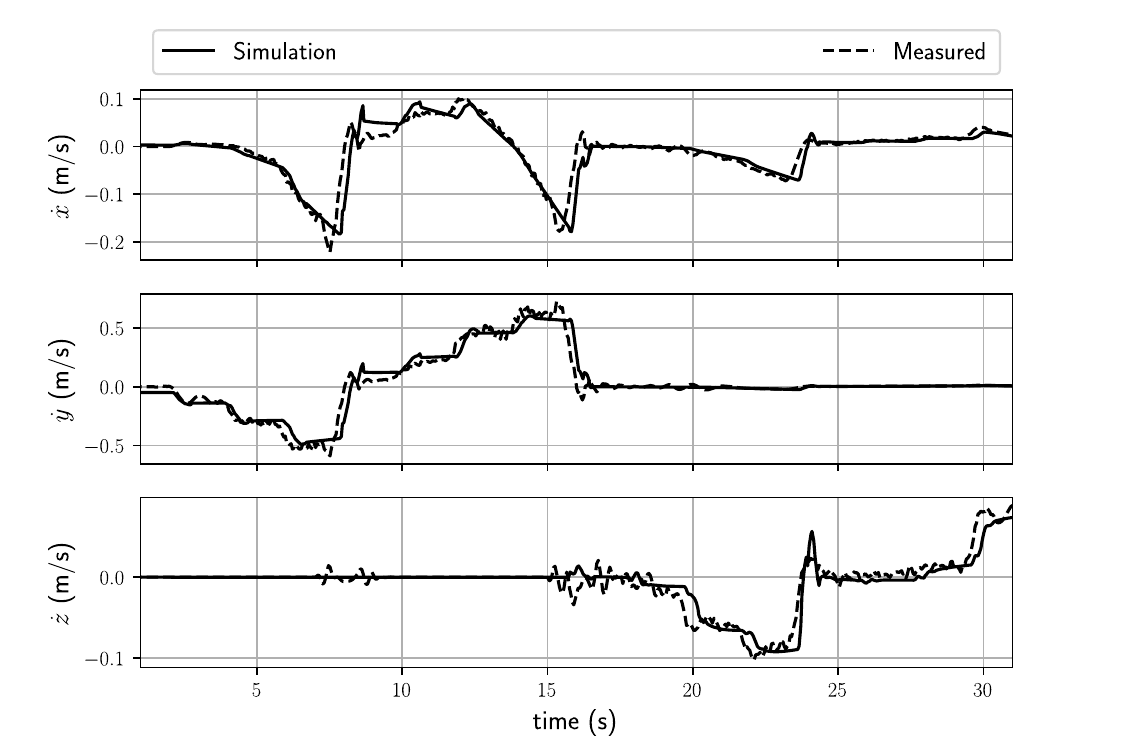}
    \caption{Velocity difference in the task space}
  \end{subfigure}
\end{minipage}%
\begin{minipage}{0.48\textwidth}
  \centering
  \begin{subfigure}{\linewidth}
    \centering
        \includegraphics[scale=0.32]{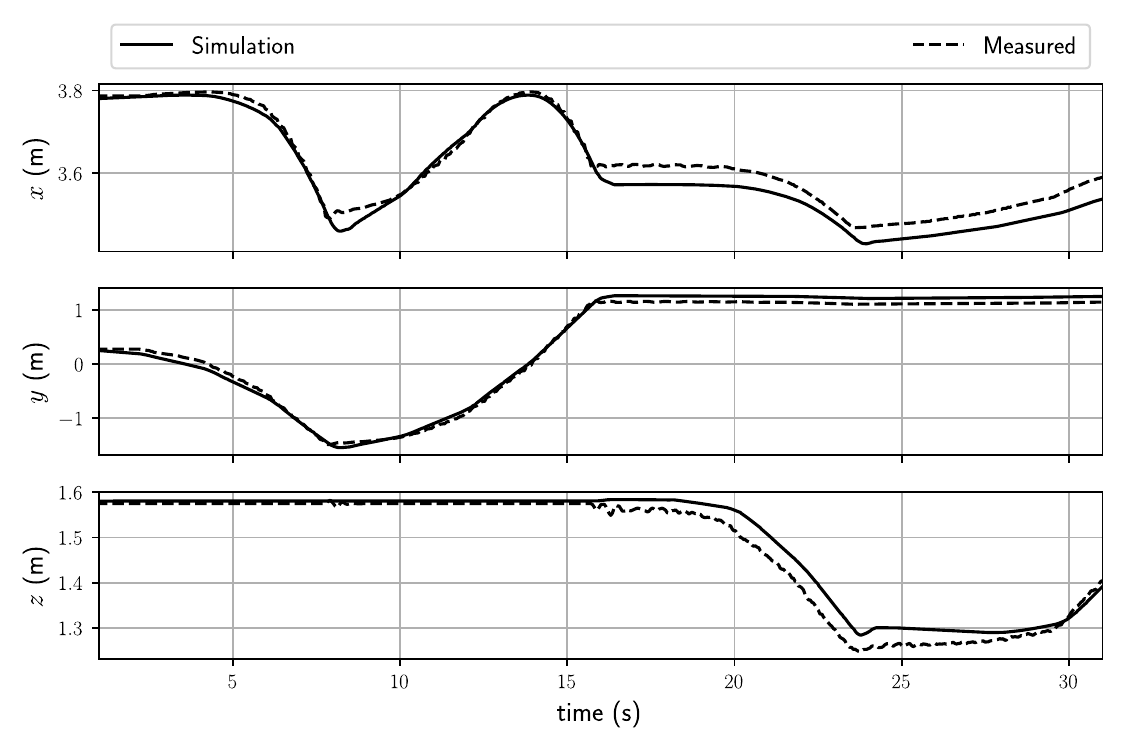}
    \caption{Position difference in the task space}
  \end{subfigure}
\end{minipage}

\begin{minipage}{0.48\textwidth}
  \centering
  \begin{subfigure}{\linewidth}
    \centering
    \includegraphics[scale=0.32]{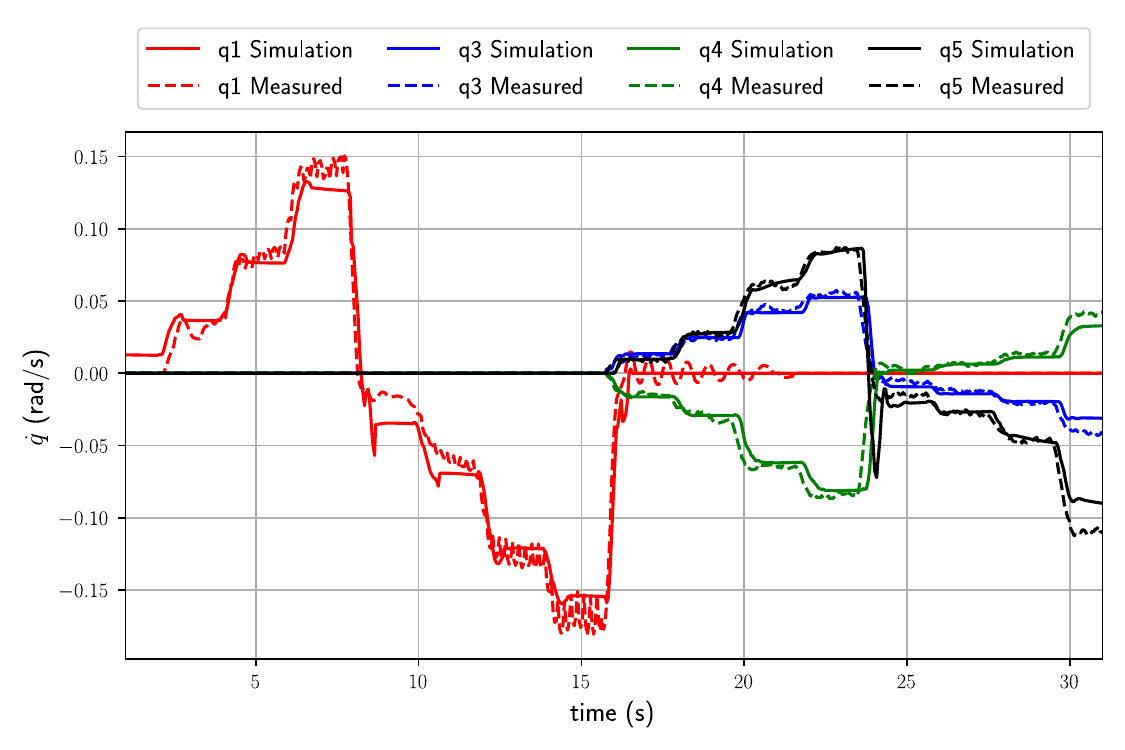}
    \caption{Velocity difference in the joint space}
  \end{subfigure}
\end{minipage}%
\begin{minipage}{0.48\textwidth}
  \centering
  \begin{subfigure}{\linewidth}
    \centering
    \includegraphics[scale=0.32]{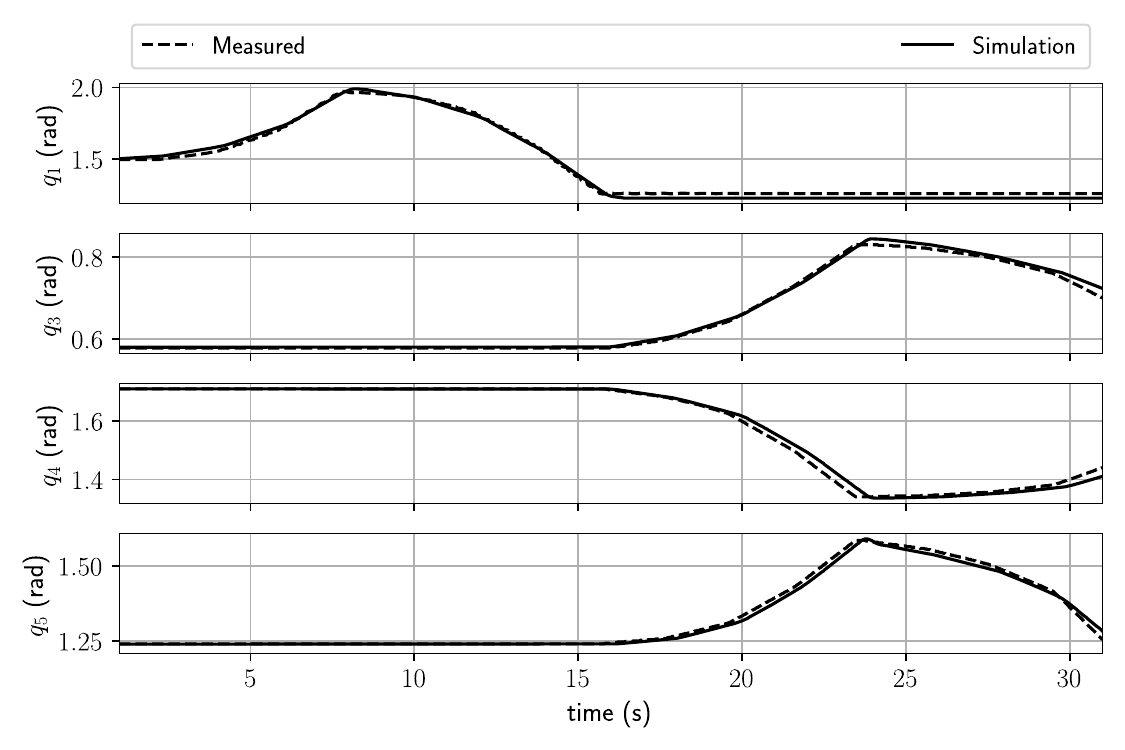}
    \caption{Position difference in the joint space}
  \end{subfigure}
\end{minipage}

\caption{Comparison between the simulated and the measured movement.}
\label{simu_real}
\end{figure}

\subsubsection{Training Procedure}
For effective training of the agent, maintaining a high sampling rate is crucial to facilitate learning the designated tasks. In this particular study, we employed the IsaacGym simulator \cite{isaacgym}, a simulation environment specifically designed to support policy learning with a parallelized physics engine on a single GPU. This parallelization enables a high sampling rate, enhancing the agent's learning capabilities. Following the completion of each episode, we randomly reset the initial arm configuration. This approach allows the trained agent to tackle the task from various starting positions, promoting its adaptability. If the maximum episode length is reached, we terminate the episode. After the random reset of the arm configuration, there are instances where the initial arm configuration intersects with the ground. In such cases, we immediately terminate the episode without providing any rewards.

For training the control policy in task space control, the Proximal Policy Optimization (PPO) algorithm \cite{ppo} is employed. Specifically, the implementation of PPO provided by Makoviichuk et al. \cite{rlgames} is utilized, which supports GPU-accelerated training with parallel environments using Isaac GYM. To approximate the value function and policy, two multilayer perceptron (MLP) neural networks are employed. The hidden layers of these networks utilize the Exponential Linear Unit (ELU) activation function, while the output layer is linear. The network architecture consists of three hidden layers, with 256 units in the first layer, 128 units in the second layer, and 64 units in the final layer. The control policy is trained at a frequency of 20 Hz, utilizing a computer equipped with an Intel Core i7-9850H CPU, 16GB of RAM, and an NVIDIA Quadro RTX 3000 GPU. To expedite the training process, 256 agents are employed in parallel, as shown in Figure \ref{isaac}. The training process takes around 2.5 hours to converge.

\section{Experimental Evaluation}
The objective of this study is to develop a task space control policy in simulation that can be directly deployed on a real machine without any modifications. This requires minimizing the gap between the simulation and the real-world performance (referred to as the Sim2Real Gap). To achieve this, we first assess the effectiveness of a data-driven actuator model integrated into the simulation. Subsequently, we demonstrate the performance of the trained task space control policy on a real-world Brokk 170 machine.

\subsection{Sim2Real Gap}
The purpose of this experiment is to evaluate the disparity between the behavior of a simulated machine and a real machine, both in terms of task space and joint space. To achieve this, Brokk 170 was initially set to a specific pose defined by the joint configuration vector $\textbf{q}_s = [1.5, 0.5, 0.58, 1.71, 1.24]$, see Fig. \ref{simu_real}a. A predetermined sequence of control signals, spanning a duration of 30 seconds, was applied to the joints of the robot. This control signal sequence was intentionally designed to induce diverse changes in joint angles and simultaneous actuation of multiple joints. In order to assess the performance of the simulation, the same control signal sequence was also applied to the simulated Brokk 170. The control signals from the predefined sequence were applied at a frequency of 20 Hz throughout the experiments. Similarly, the joint angle and velocity values, as well as the end-effector position and velocity, were recorded at the same frequency of 20 Hz.

The experimental results are depicted in Figure \ref{simu_real}. Notably, in the simulation, the error accumulated over time due to the reliance on predicted joint angle values for subsequent predictions. However, despite this accumulation, the discrepancy between the simulation and the real machine, both in the joint and task space, remained reasonably small. This outcome suggests that the integrated data-driven actuator model effectively captures the underlying nonlinear effects of the hydraulic system.

\subsection{Task Space Control}
In this experiment, we assess the effectiveness of the trained policy for task space control by directly deploying it onto the Brokk 170 machine without any additional tuning. The control policy is designed to generate control signals based on the desired goal velocity in task space. It operates in the background at a frequency of 20 Hz and communicates with the machine control unit (MCU) responsible for controlling the actuators. To evaluate the performance of the trained policy, we compare it with a well-established method based on the pseudo-inverse Jacobian matrix \cite{Khatib}. 

\begin{equation}
\begin{gathered}
    \dot{\textbf{q}}_t = \textbf{J}^+ (\textbf{q}_t)(\textbf{x}_{goal}+ \textbf{K}\textbf{e}_t)
\end{gathered}
\end{equation}

,where $\textbf{J}^+$ represents the pseudo-inverse of the Jacobian \cite{buss}, $\textbf{K} \in \mathbb{R}^{3x3}$ is a positive definite gain matrix and $\textbf{e} \in \mathbb{R}^{3x1}$ is the remained velocity error. In this work, $\textbf{J}^+$ is defined by introducing the weighted least squares method \cite{chan} and damped least square method \cite{nakamura} to handle the joint limit and singularity problem, respectively: 

\begin{equation}
\begin{gathered}
    \textbf{J}^+ = \textbf{W}^{-1}\textbf{J}^T(\textbf{J}\textbf{W}\textbf{J}^T+\lambda^2\textbf{I})^{-1}
\end{gathered}
\end{equation}
,where the diagonal matrix $\textbf{W} \in \mathbb{R}^{5x5}$ is utilized to penalize the joint motion when a joint approaches its hardware limit. The elements of $\textbf{W}$ are adjusted to impose suitable penalties based on the proximity of each joint to its limit. Additionally, the damping factor $\lambda$ is employed to restrict the robot's motion when it approaches a singularity, preventing undesired behavior.

\begin{figure}[H]
\begin{minipage}{1\textwidth}
  \centering
  \begin{subfigure}{\linewidth}
    \centering
    \includegraphics[scale=0.45]{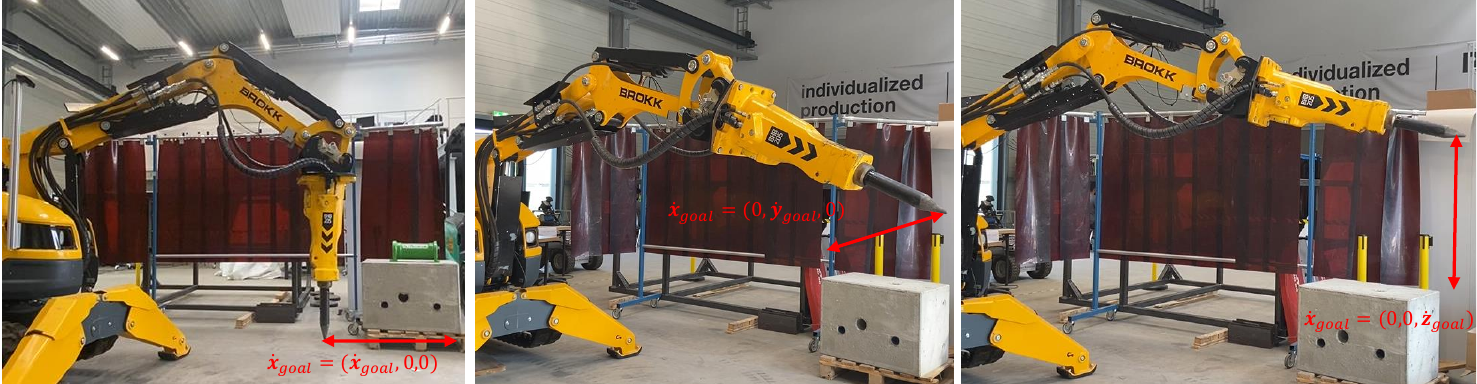}
    \caption{Different start configuration for the task space control test.}
  \end{subfigure}
\end{minipage}%

\centering
\begin{minipage}{0.5\textwidth}
  \centering
  \begin{subfigure}{\linewidth}
    \centering
    \includegraphics[scale=0.3]{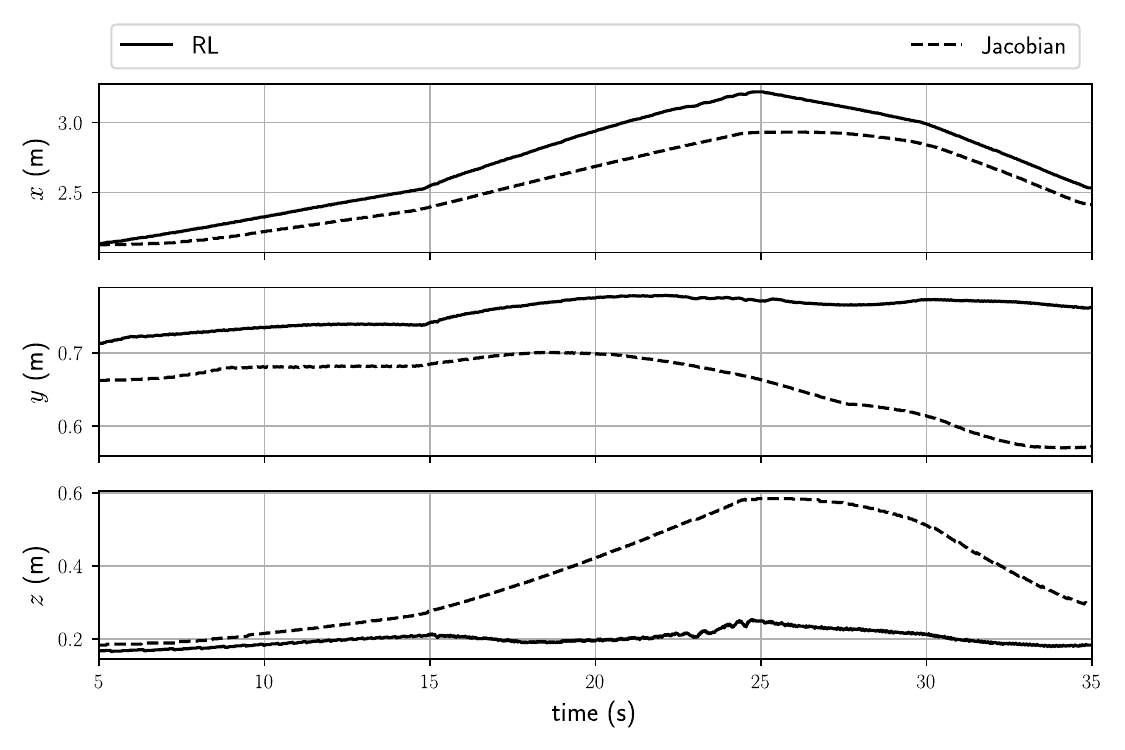}
    \caption{Position profile in the task space}
  \end{subfigure}
\end{minipage}%
\begin{minipage}{0.5\textwidth}
  \centering
  \begin{subfigure}{\linewidth}
    \centering
\includegraphics[scale=0.3]{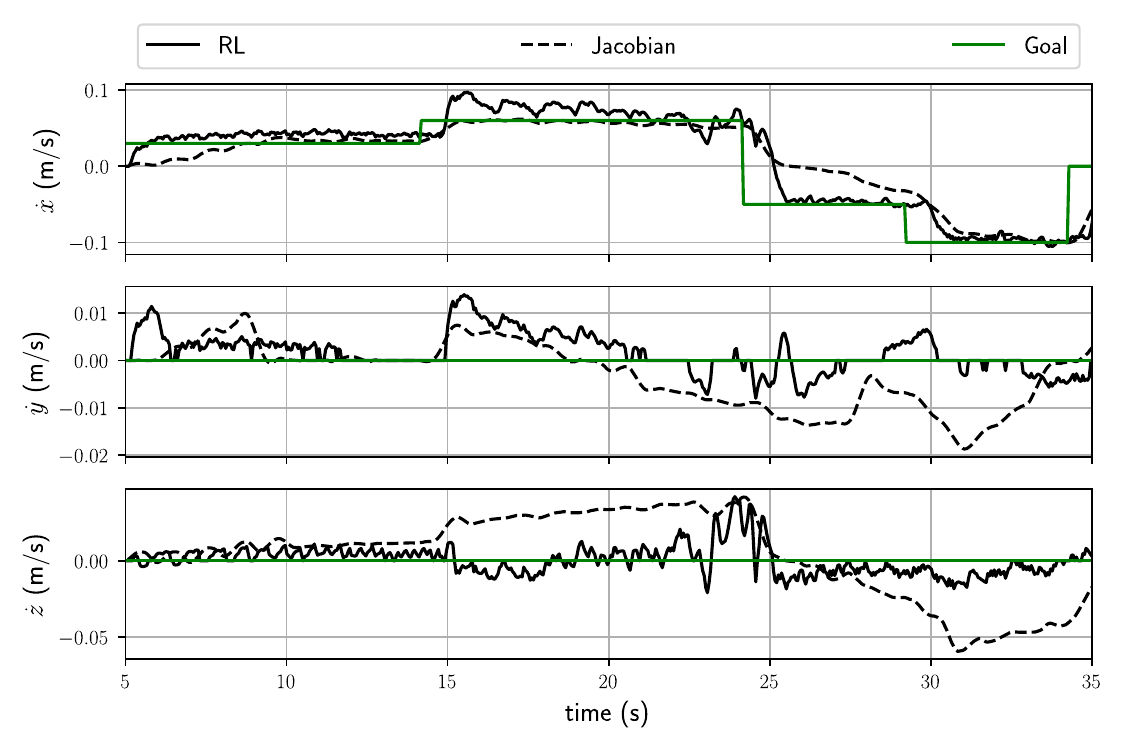}
    \caption{Velocity profile in the task space}
  \end{subfigure}
\end{minipage}

\begin{minipage}{0.5\textwidth}
  \centering
  \begin{subfigure}{\linewidth}
    \centering
    \includegraphics[scale=0.3]{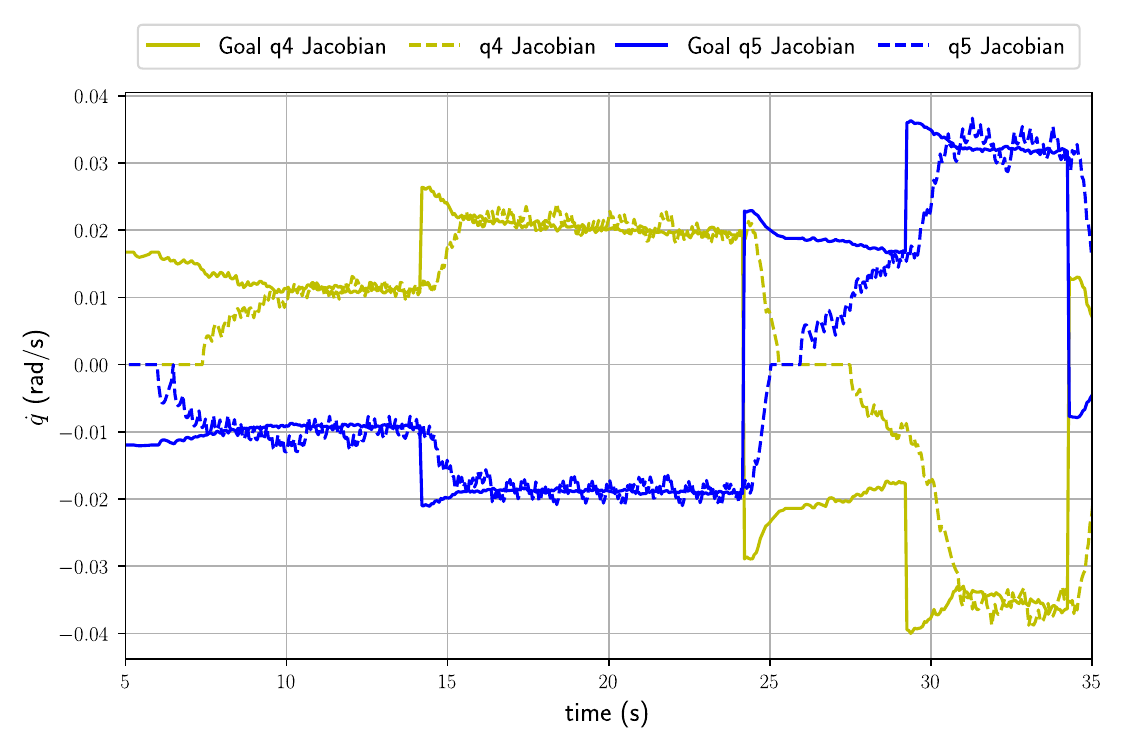}
    \caption{Velocity profile in the joint space}
  \end{subfigure}
\end{minipage}

\begin{minipage}{0.5\textwidth}
  \centering
  \begin{subfigure}{\linewidth}
    \centering
        \includegraphics[scale=0.3]{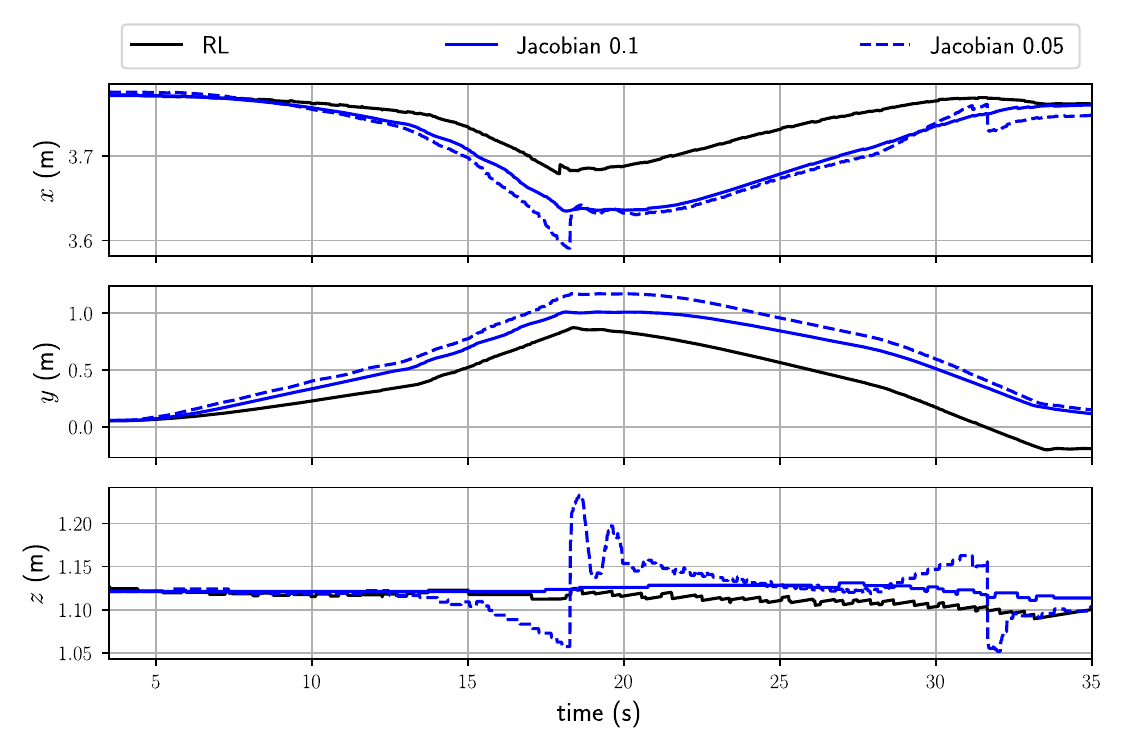}
    \caption{Position profile in the task space}
  \end{subfigure}
\end{minipage}%
\begin{minipage}{0.5\textwidth}
  \centering
  \begin{subfigure}{\linewidth}
    \centering
            \includegraphics[scale=0.3]{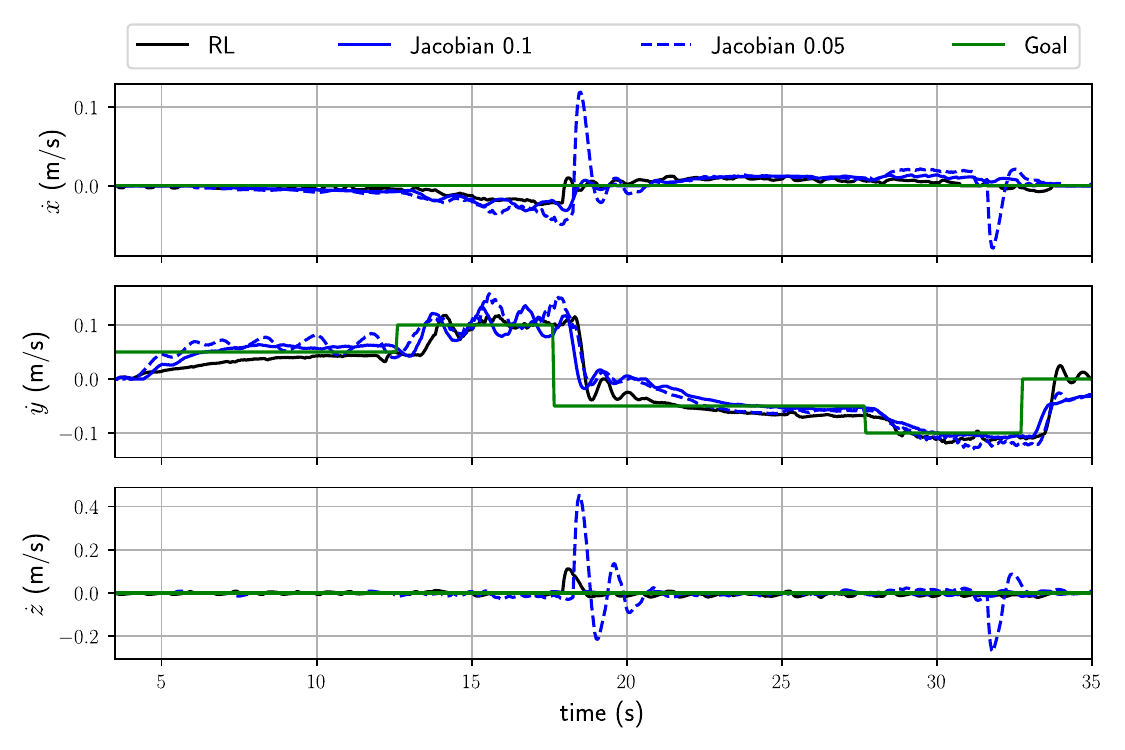}
    \caption{Velocity profile in the task space}
  \end{subfigure}
\end{minipage}

\begin{minipage}{0.5\textwidth}
  \centering
  \begin{subfigure}{\linewidth}
    \centering
                \includegraphics[scale=0.3]{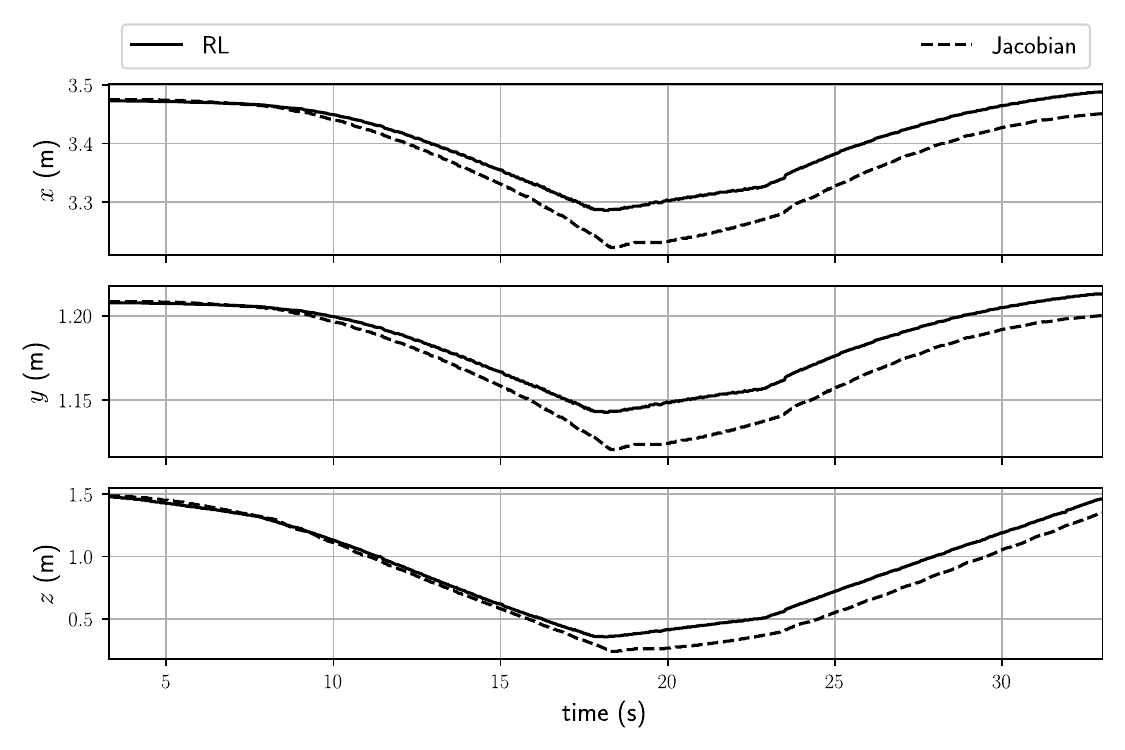}
    \caption{Position profile in the task space}
  \end{subfigure}
\end{minipage}%
\begin{minipage}{0.5\textwidth}
  \centering
  \begin{subfigure}{\linewidth}
    \centering
                    \includegraphics[scale=0.3]{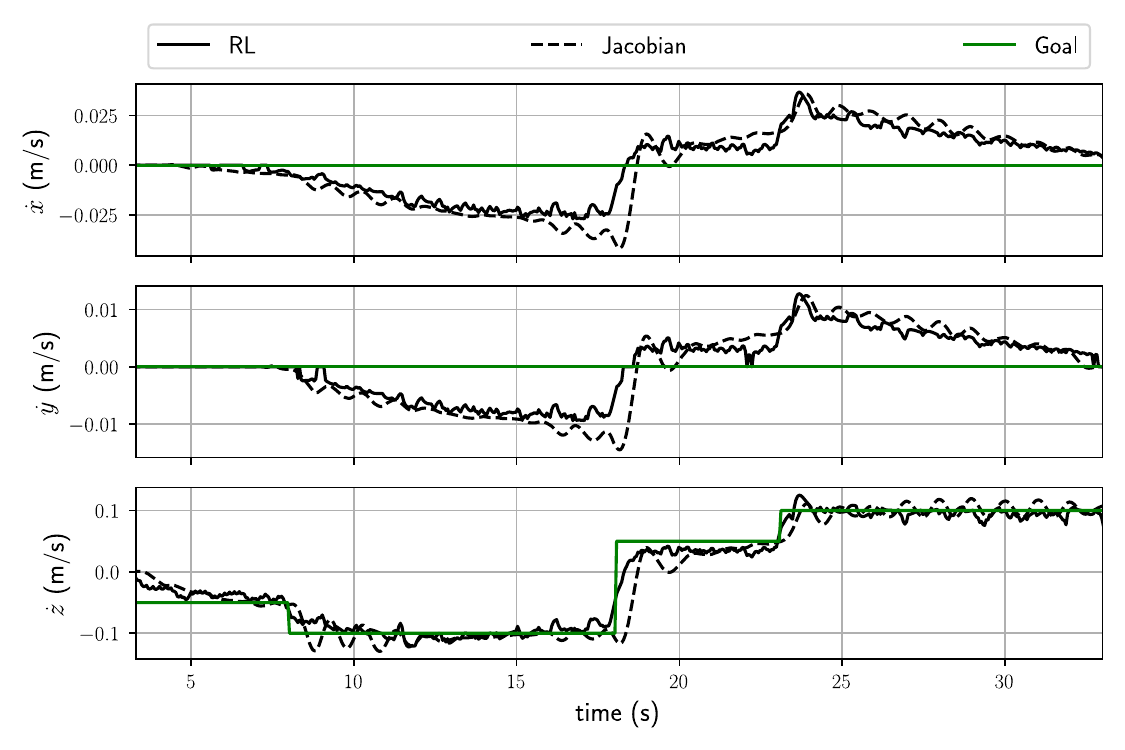}
    \caption{Velocity profile in the task space}
  \end{subfigure}
\end{minipage}

\caption{Results during $x$-direction (b-d), $y$-direction (e,f) and $z$-direction (g,h).}
\label{task_space_control}
\end{figure}

In order to assess and contrast these two distinct approaches under identical conditions, a predetermined velocity sequence in the task space was applied to both the trained control policy and the Jacobian-based controller. This experiment encompassed three distinct velocity sequences: one in the $x$-direction, one in the $y$-direction, and one in the $z$-direction, as illustrated in Fig. \ref{task_space_control}. For the case of movement in the $x$-direction, the velocities of $y$ and $z$ were set to zero (Fig. \ref{task_space_control}c), ensuring that only the $x$ position would experience an increment while the $y$ and $z$ positions remained stationary. 

The results depicted in Figure \ref{task_space_control}b demonstrate this expected behavior of the RL agent, where the $y$ and $z$ positions remain stationary. In contrast, the Jacobian-based method significantly fails to maintain the stationary positions of $y$ and $z$, as evident from the figure. This discrepancy can be attributed to the observations presented in Figure \ref{task_space_control}d. After Equation (3) computes the desired joint motions to move the end-effector according to $\textbf{x}_goal$, the resulting $\dot{\textbf{q}}_t$ is converted into PWM signals that actuate the joints. Various advanced control methods, such as backstepping-based adaptive controllers \cite{c7}, can be utilized for this low-level control task. However, these controllers typically rely on accurate dynamic models of the system. The performance accuracy of such controllers heavily depends on the precision of the model employed. Consequently, in practice, PID controllers are often employed for this low-level control task, as in the present work. However, due to the nonlinearity of the hydraulic system, the fine-tuned gains of the PID controller yield varying results depending on the direction of movement and the velocity amplitude. Figure \ref{task_space_control}d clearly depicts how the PID gains perform well when $q_4$ moves in the positive direction. However, around $t=25$ s, when the direction changes, the performance deteriorates.

In contrast, the trained RL agent directly outputs control signals in the form of PWM signals, as shown in Figure \ref{overview}. During the training process, the RL agent explores and learns the relationships between control signals and resulting end-effector velocities, enabling robust velocity tracking, as illustrated in Figures 3c, 3f, and 3h.

As previously mentioned, the damped least square method is employed to address the singularity problem, aiming to dampen the robot's motion using a damping factor $\lambda$ when it approaches a singularity. Typically, $\lambda$ is determined based on a constant $k$ and the current manipulator configuration \cite{nakamura}. However, this damping effect can impede accurate task space control. By reducing the damping effect through a lower value of the constant $k$, the system often exhibits jerky movements near the singularity, particularly when the arm is almost fully extended, as illustrated in Figure \ref{task_space_control}a. The resulting jerky movements are also evident in the position and velocity profiles, as shown in Figures \ref{task_space_control}e and \ref{task_space_control}f at $t=18$ s when $k$ is reduced from 0.1 to 0.05. In contrast, the RL agent demonstrates a more robust performance, which can also be verified in case of movement in $z$-direction, see Fig. \ref{task_space_control}g and \ref{task_space_control}h.

\section{Conclusion}
This paper introduces a reinforcement learning (RL) framework designed to learn an effective policy for task space control in a simulation environment. The goal is to enable direct deployment of the learned policy to a real construction machine without any modifications. To bridge the gap between simulation and reality, a data-driven actuator model is incorporated during training to capture the machine-specific nonlinearities in the relationship between control inputs and system state changes. The learned control policy takes the desired velocities in the $x$-, $y$-, and $z$-directions in task space as input and directly generates the corresponding control signals. Compared to conventional methods, the RL-based approach has the advantage of not relying on a dynamical model, making it suitable for hydraulic machines where such models are typically unavailable. Additionally, the proposed method outperforms the Jacobian-based approach by eliminating damping effects and the need for a low-level controller.

In the context of teleoperation, an assistance system is proposed, which enables intuitive task space control. This system enhances safety during teleoperation and minimizes errors during task execution, particularly for heavy-duty hydraulic machines. The effectiveness of the proposed method is evaluated through experiments conducted on a full-scale hydraulic machine, Brokk 170. In future work, the authors plan to further investigate the impact of the proposed method on mental workload and task efficiency by conducting additional explorations and evaluations involving a larger number of participants.

\subsubsection{Acknowledgement} 
This work has been supported by the North Rhine-Westphalia Ministry of Economic Affairs, Innovation, Digitalisation and Energy of the Federal Republic of Germany under the research intent 5G.NAMICO: Networked, Adaptive Mining and Construction.
%
%

\end{document}